\documentclass[lettersize,journal]{IEEEtran}

\usepackage{amsmath,amsfonts,amssymb}
\usepackage{mathtools}
\usepackage{mathrsfs}
\usepackage{bbm}
\usepackage{bbold}
\usepackage{algorithmic}
\usepackage{algorithm}

\usepackage{array}
\usepackage{booktabs}
\usepackage{multirow}
\usepackage{diagbox}
\usepackage{graphicx}
\usepackage[
    caption=false,
    font=normalsize,
    labelfont=sf,
    textfont=sf
]{subfig}

\usepackage{textcomp}
\usepackage{xcolor}
\usepackage{comment}
\usepackage[shortlabels]{enumitem}
\usepackage{verbatim}
\usepackage{stfloats}
\usepackage{placeins}
\usepackage{balance}
\usepackage{url}
\usepackage{xr}
\usepackage{cite}

\newtheorem{theorem}{Theorem}

\def\BibTeX{{\rm B\kern-.05em{\sc i\kern-.025em b}\kern-.08em
    T\kern-.1667em\lower.7ex\hbox{E}\kern-.125emX}}

\usepackage{etoolbox}
\makeatletter
\patchcmd{\section}
  {3.0ex plus 1.5ex minus 1.5ex}
  {1.9ex plus 0.4ex minus 0.4ex}
  {}{\PackageError{compactlayout}{Cannot patch section spacing}{Check IEEEtran version.}}
\patchcmd{\section}
  {0.7ex plus 1ex minus 0ex}
  {0.5ex plus 0.2ex minus 0ex}
  {}{\PackageError{compactlayout}{Cannot patch section afterskip}{Check IEEEtran version.}}
\patchcmd{\subsection}
  {3.5ex plus 1.5ex minus 1.5ex}
  {1.9ex plus 0.4ex minus 0.4ex}
  {}{\PackageError{compactlayout}{Cannot patch subsection spacing}{Check IEEEtran version.}}
\patchcmd{\subsection}
  {0.7ex plus .5ex minus 0ex}
  {0.5ex plus 0.2ex minus 0ex}
  {}{\PackageError{compactlayout}{Cannot patch subsection afterskip}{Check IEEEtran version.}}
\makeatother

\setlist[itemize]{topsep=3pt plus 1pt minus 1pt,
  itemsep=1pt plus 0.5pt minus 0.5pt,parsep=0pt,partopsep=0pt}
\setlist[enumerate]{topsep=3pt plus 1pt minus 1pt,
  itemsep=1pt plus 0.5pt minus 0.5pt,parsep=0pt,partopsep=0pt}

\AtBeginDocument{%
  \setlength{\parskip}{0pt}
  \setlength{\abovedisplayskip}{4pt plus 1pt minus 1pt}%
  \setlength{\belowdisplayskip}{4pt plus 1pt minus 1pt}%
  \setlength{\abovedisplayshortskip}{0pt plus 1pt}%
  \setlength{\belowdisplayshortskip}{3pt plus 1pt minus 1pt}%
  \setlength{\jot}{2pt}
  \setlength{\textfloatsep}{5pt plus 1pt minus 1pt}%
  \setlength{\dbltextfloatsep}{5pt plus 1pt minus 1pt}%
  \setlength{\floatsep}{5pt plus 1pt minus 1pt}%
  \setlength{\dblfloatsep}{5pt plus 1pt minus 1pt}%
  \setlength{\intextsep}{5pt plus 1pt minus 1pt}%
  \setlength{\abovecaptionskip}{3.5pt}%
  \setlength{\belowcaptionskip}{0pt}%
  \raggedbottom 
}

\begin{document}

\title{Shape-Adaptive Subspace-Constrained Quadratic Matrix Factorization}

\author{Zheng~Zhai, Zhuan~Liang and Sujun~Guo%
\thanks{This work was supported by the National Natural Science
Foundation of China (NSFC) under Grant 12301478.}%
\thanks{The authors are with the Department of Statistics,
Faculty of Arts and Sciences, Beijing Normal University,
Zhuhai, China.  }%
\thanks{Corresponding author: Zheng Zhai. E-mail: zhaizheng@bnu.edu.cn.}
}

\markboth{}{}

\maketitle

\begin{abstract}

Quadratic manifold fitting provides a flexible framework for modeling nonlinear low-dimensional structures embedded in high-dimensional spaces. However, existing methods typically employ prescribed noise models or fixed loss functions, limiting their adaptability to varying noise characteristics. To address this limitation, we propose an alternating maximum likelihood estimation method for subspace-constrained quadratic matrix factorization, termed AMLE-SQMF, which jointly estimates the quadratic manifold and the underlying noise distribution. Specifically, we model the residuals using a generalized Gaussian distribution (GGD), whose shape parameter controls the tail behavior and determines the associated fitting loss. By exploiting the closed-form maximum likelihood estimator of the scale parameter, we eliminate the scale variable and derive a profile negative log-likelihood. Based on this formulation, we develop an alternating optimization algorithm that iteratively updates the manifold parameters and estimates the noise shape parameter. This enables adaptive loss selection within a unified likelihood-based framework, without requiring prior specification of the noise characteristics. Experiments on synthetic and real-world datasets demonstrate the effectiveness of AMLE-SQMF in quadratic manifold fitting and its robustness to diverse noise distributions.

\end{abstract}

\begin{IEEEkeywords}
adaptive, quadratic, subspace-constrained optimization, manifold learning
\end{IEEEkeywords}

\section{Introduction}
\label{sec:introduction}

High-dimensional data often exhibit low-dimensional structure characterized by a nonlinear manifold. Learning this structure is a fundamental manifold-learning problem with applications in dimensionality reduction~\cite{zhang2004principal}, reconstruction, denoising~\cite{fefferman2018fitting}, and geometric data analysis~\cite{Aamari}. When only finitely many noisy observations of an unknown manifold are available, the challenge is to capture its local geometry with a computationally tractable model that is robust to noise.

A natural approach uses a local low-order polynomial approximation. A quadratic model captures both linear structure and second-order geometry, including curvature, and can improve on linear approximation while retaining a simple parametric form. Suitable subspace constraints explicitly characterize the local center, tangent space, normal space, and second fundamental form~\cite{docarmo2016differential}, motivating subspace-constrained quadratic functions~\cite{zhai2025subspace} for local manifold approximation.

The accuracy of quadratic manifold approximation depends on parameter estimation. Existing methods commonly use squared Euclidean or Frobenius-norm losses corresponding to Gaussian noise assumptions. These convenient objectives may be sensitive to outliers and inadequately describe non-Gaussian residuals~\cite{huber1964robust}. Residuals can include observation noise and the approximation error from second-order truncation, so their distribution need not be Gaussian. A fixed loss can therefore limit adaptation to different data characteristics.

To address this limitation, we develop an adaptive subspace-constrained quadratic approximation framework based on maximum likelihood estimation~\cite{casella2002statistical}. Rather than specifying a fixed distribution for the observation noise alone, we model the residuals of the quadratic approximation using a generalized Gaussian distribution with an unknown shape parameter~\cite{roenko2014estimation,pascal2013parameter}. This formulation provides a flexible family of residual models with different tail behaviors. In particular, the Gaussian model is recovered when the shape parameter equals two, whereas the Laplace model corresponds to a shape parameter of one~\cite{roenko2014estimation}. Estimating the shape parameter from the data allows the fitting procedure to adapt its residual model to the observed data, instead of relying on a prespecified noise distribution.

The estimation problem jointly determines the quadratic model parameters, latent coordinates, and generalized Gaussian shape parameter. Eliminating the scale through its maximum likelihood estimate yields a profile likelihood. We alternate updates of the model and shape parameters while explicitly preserving the orthonormality and mutual orthogonality of the tangent and normal bases~\cite{zhai2025subspace}. This integrates adaptive residual modeling with geometrically interpretable quadratic approximation.

The main contributions of this work are summarized as follows:
The main contributions of this work are summarized as follows:
\begin{itemize}
    \item \textbf{Adaptive generalized Gaussian noise modeling.}
    We establish a maximum likelihood framework based on the generalized Gaussian distribution, in which the shape parameter is jointly estimated with the manifold model. By analytically profiling out the scale parameter, we derive a reduced likelihood objective that enables adaptive modeling of diverse residual distributions.

    \item \textbf{Alternating maximum likelihood estimation.}
    We propose an alternating optimization algorithm to jointly estimate the manifold parameters and noise shape parameter while preserving the orthogonality constraints of the subspace bases.

    \item \textbf{Comprehensive numerical validation.}
    We conduct experiments on synthetic and real-world datasets to evaluate the approximation performance and robustness of the proposed method under different noise conditions. We further investigate the effects of ambient dimension, sample size, and noise characteristics on shape parameter estimation and manifold approximation.
\end{itemize}

The remainder of this paper is organized as follows. Section II introduces the adaptive subspace-constrained quadratic manifold model and formulates the generalized Gaussian residual distribution. Section III develops the joint maximum likelihood estimation framework, including the profile likelihood and the alternating estimation procedure, and establishes its convergence properties. Section IV presents the geometric optimization formulation, derives the corresponding gradients, and analyzes the orthogonality constraints and first-order optimality conditions. Section V develops efficient optimization algorithms, including Riemannian gradient descent and the proposed immediate residual gradient update strategy. Section VI reports numerical experiments to evaluate computational efficiency, statistical estimation accuracy, and manifold reconstruction performance. Finally, Section VII concludes the paper.

\subsection{Related works and Motivations}

We build our work upon the quadratic matrix factorization (QMF) and the subspace-constrained quadratic matrix factorization (SQMF) formulation, which we previously developed in~\cite{zhai2024quadratic, zhai2025subspace} to learn latent manifold structures. QMF formulates quadratic factorization by allowing the parameters associated with the linear and quadratic components to be optimized independently, without imposing orthogonality or other structural constraints. In contrast, SQMF provides a geometrically structured representation of local manifold geometry by explicitly characterizing the tangent and normal subspaces, together with the second fundamental form, which captures the local curvature and describes how the manifold deviates from its tangent space along the normal directions. This geometric structure establishes a natural connection between the tangent-space representation and the corresponding normal-space variations, enabling a more expressive approximation of nonlinear manifolds.

The quadratic factorization framework is closely related to a broad class of manifold fitting and denoising methods, which aim to recover low-dimensional geometric structures embedded in high-dimensional observations, typically in the presence of noise and irregular sampling. Representative approaches include local linear fitting methods, such as local principal component analysis (LPCA)~\cite{lpca}; ridge-based methods derived from kernel density estimation (KDE)~\cite{genovese2014nonparametric} and its logarithmic variant (LOG-KDE)~\cite{ozertem2011locally}; tangent-space aggregation methods, such as manifold fitting (MFIT)~\cite{fefferman2018fitting}; and higher-order regression techniques, including moving least squares (MLS)~\cite{sober2020manifold}. Other approaches employ alternative geometric primitives to represent local manifold structures, such as spherelets, which approximate manifolds using low-dimensional spheres embedded in high-dimensional ambient spaces~\cite{li2022efficient}. Collectively, these methods offer complementary perspectives on manifold approximation through local linearization, implicit geometric estimation, and higher-order or geometrically structured local representations.

Despite their differences in geometric representations, including density ridges, tangent spaces, spherelets, and polynomial approximations, these methods commonly rely on residual-based objectives to quantify the discrepancy between the estimated manifold and the underlying data structure. In particular, the residual measures the deviation of an observation from its corresponding estimated point on the manifold. The squared Euclidean loss is widely adopted in these methods, largely due to its analytical tractability and computational convenience. However, its statistical justification is closely associated with the Gaussian noise assumption, which may be inadequate in practical settings involving heavy-tailed noise or outliers. Moreover, the residual generally reflects not only observational noise but also manifold approximation and estimation errors. Consequently, its distribution may deviate substantially from Gaussianity, potentially compromising the robustness of conventional least-squares-based methods.

To address this limitation, we consider the generalized Gaussian distribution (GGD)~\cite{varanasi1989parametric} as a more flexible probabilistic model for the residuals. Its shape parameter \(p\) controls the tail behavior of the distribution, allowing it to accommodate a broader range of noise characteristics than the conventional Gaussian model. However, selecting an appropriate value of \(p\) is itself a nontrivial problem, as the underlying residual distribution is generally unknown and may vary across datasets. Rather than prescribing \(p\) in advance or relying on heuristic parameter selection, we develop an adaptive maximum likelihood estimation framework that jointly estimates the GGD shape parameter \(p\) and the parameters of the quadratic manifold model. By integrating robust probabilistic modeling with SQMF, the proposed approach simultaneously adapts the geometric representation to the underlying manifold structure and the residual distribution to the observed data, providing a unified framework for robust manifold approximation and denoising.

\section{Formulation of Adaptive subspace-constrained quadratic approximation model}
\label{sec:adaptive_formulation}

To model the underlying low-dimensional structure of the observed data, we formulate the approximation problem as
\begin{equation}
\label{eq:model_general}
x_i=f(\tau_i)+\epsilon_i,\qquad i=1,\ldots,n,
\end{equation}
where $\tau_i$ denotes the low-dimensional coordinate of $x_i$, $f$ is a smooth mapping that locally parameterizes the underlying manifold, and $\epsilon_i$ stands for the observation noise. This representation is geometrically justified by the local chart structure of a smooth manifold, which guarantees the existence of local coordinates and a smooth parameterization in a neighborhood of each point. 

Since the space of smooth functions is infinite-dimensional, directly seeking an optimal mapping $f$ is generally intractable. We therefore restrict $f$ to a finite-dimensional parametric family of subspace-constrained quadratic functions. Beyond improving tractability, this formulation provides an explicit geometric interpretation of the local manifold structure by characterizing its center, tangent space, normal space, and second fundamental form. The class of subspace-constrained quadratic functions is defined as follows:
\begin{equation}\notag
\begin{aligned}
\mathcal{F}=\bigl\{f:\mathbb{R}^d\to\mathbb{R}^D\;\big|\;&
f(\tau)=c+U\tau+V\mathcal{A}(\tau,\tau),\\
&U^\top U=I_d,\quad V^\top V=I_s,\\
&U^\top V=0\bigr\}.
\end{aligned}
\end{equation}
where $c\in\mathbb{R}^D$ is the center parameter,
$U\in\mathbb{R}^{D\times d}$ is an orthonormal basis for the local
tangent subspace, and $V\in\mathbb{R}^{D\times s}$ is an orthonormal
basis for the subspace associated with the quadratic components. The bilinear map $\mathcal{A}:\mathbb{R}^d\times\mathbb{R}^d\to\mathbb{R}^s$ is defined as the quadratic form on the $k$-th slice of $\mathcal A$, i.e.,
\begin{equation}
\label{eq:sqmf_tensor_representation}
[\mathcal{A}(\tau_i,\tau_i)]_k
=
\sum_{a=1}^{d}\sum_{b=1}^{d}
\mathcal{A}_{kab}\tau_{ai}\tau_{bi}.
\end{equation}

The orthogonality constraints ensure that the tangent and normal bases are mutually orthogonal, allowing the linear and quadratic terms to capture the first- and second-order geometric structure of the manifold, respectively.

Suppose $\widehat{f}$ as the best approximation of $f$ in $\cal F$, the residual associated with the quadratic approximation satisfies
\[
r_i=x_i-\widehat{f}(\tau_i)
=\bigl(f(\tau_i)-\widehat{f}(\widehat{\tau}_i)\bigr)+\epsilon_i,
\]
where the first term denotes the approximation error and the second term represents the noise. Assuming sufficient smoothness of \(f\), the approximation error satisfies
\[
f(\tau_i)-\widehat{f}(\widehat{\tau}_i)
=O\bigl(\max\{\|\tau_i\|_2^3,\|\widehat{\tau}_i\|_2^3\}\bigr),
\]
which represents the higher-order geometric information that cannot be captured by the quadratic model.

Since the residuals $r_i$ incorporate both the approximation error and the observation noise, we model them directly using a generalized Gaussian distribution with an unknown shape parameter $p$, rather than imposing a specific prior distribution on $\epsilon_i$. The shape parameter $p$ controls the tail behavior of the distribution: a larger $p$ generally yields lighter tails, whereas a smaller $p$ produces heavier tails. Thus, estimating $p$ from the data allows the model to adapt to different residual distributions without prescribing their tail behavior in advance. In other words, we assume
\begin{equation}\label{eq:generalized_gaussian}
p(r_i\mid p,\eta)
=
\frac{1}{Z_{D,p,\eta}}
\exp\left(-\frac{\|r_i\|_p^p}{\eta}\right),\quad \eta>0,
\end{equation}
where \(Z_{D,p,\eta}\) is the normalizing constant. Here, the exponent \(p\) controls the tail behavior of the noise distribution, allowing the model to accommodate a continuum of noise assumptions between the multivariate Laplace-type model at \(p=1\) and the isotropic Gaussian model at \(p=2\). 

When \(p\) and \(\eta\) are known, maximum likelihood estimation (MLE)~\cite{casella2002statistical} of the manifold mapping and latent coordinates is equivalent, up to an additive constant, to minimizing the negative log-likelihood:
\begin{equation}\label{eq:mle_general}
\min_{f\in\mathcal{F},\,\{\tau_i\}_{i=1}^n}
\sum_{i=1}^n
\|x_i-f(\tau_i)\|_p^p.
\end{equation}
Consequently, the choice of \(p\) determines the residual penalty and directly affects the estimated manifold geometry. 
However, the noise characteristics are generally unknown in practice, and prescribing \(p\) in advance may result in a suboptimal approximation.

To address this limitation, we propose an \emph{adaptive subspace-constrained quadratic approximation model} in which the shape parameter \(p\) is estimated jointly with the manifold parameters and latent coordinates. Specifically, we consider
\begin{align}
\min_{\substack{
f\in\mathcal{F},\;\{\tau_i\}_{i=1}^n,\\ 
p>0,\;\eta>0
}}
\quad
&-\sum_{i=1}^n\log p(r_i\mid p,\eta) \label{eq:adaptive_mle}\\ 
={}&
\frac{1}{\eta}\sum_{i=1}^n
\|x_i-f(\tau_i)\|_p^p
+n\log Z_{D,p,\eta}. \notag
\end{align}
The normalizing term \(n\log Z_{D,p,\eta}\) is essential when estimating \(p\) by maximum likelihood and cannot be discarded as a constant because it depends on the unknown distribution parameters. This joint formulation allows the model to adapt its residual penalty to the observed data rather than imposing a fixed noise assumption.

The proposed model therefore unifies quadratic manifold approximation under different noise regimes within a single likelihood-based framework. By jointly adapting the manifold geometry and the noise distribution, it aims to achieve a more appropriate balance between approximation accuracy and robustness to atypical observations.

\section{Joint Maximum Likelihood Estimation of the Model Parameters and the Shape Parameter}
\label{subsec:joint_mle}

To optimize the model in~\eqref{eq:adaptive_mle}, we first derive the normalization constant \(Z_{D,p,\eta}\), which depends on the shape parameter \(p\) and scale parameter \(\eta\).


\subsubsection{Generalized Gaussian Likelihood}

The isotropic generalized Gaussian model in~\eqref{eq:generalized_gaussian} implies that residual components for each dimension are mutually independent and identically distributed (i.i.d.), each following a univariate generalized Gaussian distribution:
\[
r_{ij}\overset{\mathrm{i.i.d.}}{\sim}
\operatorname{GGD}(\alpha,p):
=
\frac{p}{2\alpha\Gamma(1/p)}
\exp\left[
-\left(\frac{|r_{ij}|}{\alpha}\right)^p
\right],
\]
where $p>0$ is the shape parameter, $\alpha>0$ is the scale
parameter, and $\Gamma(\cdot)$ is the Gamma function. This
parameterization is equivalent to~\eqref{eq:generalized_gaussian}
with $\eta=\alpha^p$. In particular,
$p=1$ corresponds to Laplace noise, whereas $p=2$ corresponds to
Gaussian noise with variance $\alpha^2/2$.

Therefore, we have the total number of residual components being $N=nD$. For notational convenience, we denote
$\Theta=(c,U,V,\mathcal{A},\{\tau_i\}_{i=1}^n).$
Under the independence assumption, the negative log-likelihood is
\[
\begin{aligned}
\mathcal{L}(\Theta,p,\alpha)
={}N\log\left(\frac{2\alpha\Gamma(1/p)}{p}\right)
+\frac{1}{\alpha^p}
\sum_{i=1}^n\sum_{j=1}^D
|r_{ij}(\Theta)|^p,
\end{aligned}
\]
where $r_{ij}(\Theta)$ denotes the $j$th component of the residual associated with the $i$th data point. Consequently, the joint MLE problem is
\begin{equation}
\label{eq:joint_mle_generalized_gaussian}
\begin{aligned}
\min_{\Theta,p,\alpha}\quad&
\mathcal{L}(\Theta,p,\alpha)\\
\text{subject to}\quad&
U^\top U=I_d,\qquad V^\top V=I_s,\\
&U^\top V=0,\qquad p>0,\quad \alpha>0.
\end{aligned}
\end{equation}

Next, we analytically eliminate the scale parameter $\alpha$, whose optimum depends on $\Theta$ and $p$. For fixed $\Theta$ and $p$, define
\[
S_p(\Theta)
=
\sum_{i=1}^n\sum_{j=1}^D
|r_{ij}(\Theta)|^p.
\]
The negative log-likelihood can then be expressed as
\[
\mathcal{L}(\Theta,p,\alpha)
=
N\log\left(
\frac{2\alpha\Gamma(1/p)}{p}
\right)
+\frac{S_p(\Theta)}{\alpha^p}.
\]
 Differentiating with respect to $\alpha$ yields
\[
\frac{\partial\mathcal{L}}{\partial\alpha}
=
\frac{N}{\alpha}
-\frac{pS_p(\Theta)}{\alpha^{p+1}}.
\]
Setting this derivative to zero gives the closed-form MLE of the scale parameter:
\begin{equation}
\label{eq:alpha_mle}
\alpha^*(\Theta,p)
=
\left(
\frac{pS_p(\Theta)}{N}
\right)^{1/p}.
\end{equation}
Substituting $\alpha^*(\Theta,p)$ into $\mathcal{L}(\Theta,p,\alpha)$ yields the profile negative log-likelihood
\begin{equation}
\label{eq:profile_mle}
\begin{aligned}
\mathcal{J}(\Theta,p)
={}&N\log\left(\frac{2\Gamma(1/p)}{p}\right)\\
&+\frac{N}{p}
\log\left(
\frac{pS_p(\Theta)}{N}
\right)
+\frac{N}{p}.
\end{aligned}
\end{equation}
Consequently, the original joint MLE problem reduces to an optimization problem over $\Theta$ and $p$:
\begin{equation}
\label{eq:reduced_joint_mle}
\min_{\Theta,\,p\in[p_{\min},p_{\max}]}
\mathcal{J}(\Theta,p),
\end{equation}
where $0<p_{\min}<p_{\max}<\infty$ are prescribed bounds on the shape parameter. Once the optimal pair $(\Theta^*,p^*)$ is obtained, the scale parameter can be recovered directly using \eqref{eq:alpha_mle}, namely,
\[
\alpha^*=\left(
\frac{p^*S_{p^*}(\Theta^*)}{N}
\right)^{1/p^*}.
\]
This reduction eliminates the need to optimize over $\alpha$ explicitly, thereby simplifying the joint estimation problem.

\subsubsection{Alternating Optimization Algorithm}
Direct joint optimization of~\eqref{eq:reduced_joint_mle} with respect to \(p\) and \(\Theta\) is challenging due to its nonlinear structure. However, fixing \(p\) reduces the problem to the nonlinear \(p\)-norm fitting problem in~\eqref{eq:mle_general}, which can be solved using gradient descent, whereas fixing \(\Theta\) yields a univariate optimization problem in \(p\). Motivated by this structure, we adopt an alternating optimization strategy, iteratively updating \(\Theta\) and \(p\) while keeping the other fixed:
\begin{algorithm}[t]
\caption{Alternating Maximum-Likelihood Estimation}
\label{alg:alternating_mle}
\begin{algorithmic}[1]
\REQUIRE Data $\{x_i\}_{i=1}^n$, latent dimension $d$,
quadratic dimension $s$, bounds $[p_{\min},p_{\max}]$,
tolerance $\varepsilon$, and maximum iteration count $K_{\max}$.
\ENSURE Estimated parameters
$\Theta^*$, $p^*$, and $\alpha^*$.

\STATE Initialize $\Theta^{(0)}$ satisfying the orthogonality
constraints, $p^{(0)}\in[p_{\min},p_{\max}]$, and $k=0$.
\REPEAT
    \STATE Update $\Theta^{(k+1)}$ by solving
    \eqref{eq:theta_update}.
    \STATE Compute the residuals
    $r_i^{(k+1)}=x_i - f_{\Theta^{(k+1)}} (\tau_i^{(k+1)})$.
    \STATE Update $p^{(k+1)}$ by minimizing
    \eqref{eq:p_update} over
    $[p_{\min},p_{\max}]$.
    \STATE Set $k\leftarrow k+1$.
\UNTIL{the stopping criterion is satisfied or
$k=K_{\max}$}
\STATE Set $\Theta^*=\Theta^{(k)}$, $p^*=p^{(k)}$,
and $\alpha^*=\alpha^*(\Theta^*,p^*)$ using~\eqref{eq:alpha_mle}.
\RETURN $\Theta^*$, $p^*$, and $\alpha^*$.
\end{algorithmic}
\end{algorithm}

Given the current estimate $p^{(k)}$, update the
manifold parameters by solving
\begin{equation}
\label{eq:theta_update}
\Theta^{(k+1)}
=
\arg\min_{\Theta\in\mathcal{C}}
\mathcal{J}(\Theta,p^{(k)}).
\end{equation}

This subproblem is precisely the $\ell_{p^{(k)}}^{p^{(k)}}$
quadratic manifold fitting problem and can be solved using a
combination of Euclidean optimization and Riemannian optimization
over the coupled tangent and normal subspaces.

Next, with $\Theta^{(k+1)}$ fixed, update the shape parameter by
\begin{equation}
\label{eq:p_update}
p^{(k+1)}
=
\arg\min_{p\in[p_{\min},p_{\max}]}
\mathcal{J}(\Theta^{(k+1)},p).
\end{equation}

The procedure alternates between fitting the quadratic manifold under the current residual loss and adapting the loss shape to the current residual distribution. 

\subsubsection{Convergence Property} 

To establish the convergence properties of the proposed alternating optimization algorithm, we first show that the negative log-likelihood is bounded below. We then prove that each alternating update yields a nonincreasing objective value. It follows from the monotone convergence theorem~\cite{rudin1976principles} that the resulting sequence of objective values converges to a finite limit.
\begin{theorem}
\label{thm:lower_bound}
Let $p\in[p_{\min},p_{\max}]$, where
$0<p_{\min}<p_{\max}<\infty$. Suppose that, for each
$p\in[p_{\min},p_{\max}]$, the problem $\min_{\Theta\in\mathcal{M}} S_p(\Theta)$
admits an optimal solution $\Theta(p)$. Define
\[
S_{\min}:=\min_{p\in[p_{\min},p_{\max}]}S_p(\Theta(p)).
\]
Then the profile negative log-likelihood
$\mathcal{J}(\Theta(p),p)$ is uniformly bounded below; specifically,
\begin{equation}
\label{eq:lower_bound}
\mathcal{J}(\Theta(p),p)\ge C,
\end{equation}
where
\begin{equation}
\label{eq:constant_lower_bound}
\begin{aligned}
C:=\min_{p\in[p_{\min},p_{\max}]}
\bigg\{&
N\log\left(\frac{2\Gamma(1/p)}{p}\right)\\
&+\frac{N}{p}\log\!\left(\frac{pS_{\min}}{N}\right)
+\frac{N}{p}
\bigg\}>-\infty.
\end{aligned}
\end{equation}
\end{theorem}

\begin{IEEEproof}
By the definition of $S_{\min}$, we have
$S_p(\Theta(p))\ge S_{\min}>0$ for all
$p\in[p_{\min},p_{\max}]$. Hence,
\[
\begin{aligned}
\mathcal{J}(\Theta(p),p)
\ge{}&
N\log\left(\frac{2\Gamma(1/p)}{p}\right)\\
&+\frac{N}{p}\log\left(\frac{pS_{\min}}{N}\right)
+\frac{N}{p}.
\end{aligned}
\]
The right-hand side is continuous on the compact interval
$[p_{\min},p_{\max}]$ and therefore attains a finite minimum.
This establishes \eqref{eq:lower_bound}.
\end{IEEEproof}

\begin{theorem}[Convergence of Alternating Optimization]
\label{thm:objective_convergence}
Let $\mathcal{J}(\Theta,p)$ denote the profile negative log-likelihood, and consider the alternating optimization scheme
\begin{equation}
\label{eq:alternating_updates}
\begin{aligned}
\Theta^{(k+1)}
&= \arg\min_{\Theta\in\mathcal{M}}
\mathcal{J}(\Theta,p^{(k)}),\\
p^{(k+1)}
&= \arg\min_{p\in[p_{\min},p_{\max}]}
\mathcal{J}(\Theta^{(k+1)},p),
\end{aligned}
\end{equation}
where $\mathcal{M}$ is the feasible set of $\Theta$ and
$0<p_{\min}<p_{\max}<\infty$. Suppose that both subproblems admit minimizers at every iteration and that $\mathcal{J}$ is bounded below along the generated sequence. Then the objective sequence
\[
\mathcal{J}_k
:=\mathcal{J}(\Theta^{(k)},p^{(k)}),
\]
is monotonically nonincreasing and converges to a finite limit.
\end{theorem}

\begin{IEEEproof}
By the optimality of the $\Theta$-update and the $p$-update, respectively, we have
\begin{align*}
\mathcal{J}(\Theta^{(k+1)},p^{(k)})
&\leq \mathcal{J}(\Theta^{(k)},p^{(k)}),\\
\mathcal{J}(\Theta^{(k+1)},p^{(k+1)})
&\leq \mathcal{J}(\Theta^{(k+1)},p^{(k)}).
\end{align*}
Consequently,
\[
\mathcal{J}_{k+1}\leq \mathcal{J}_k.
\]
Thus, $\{\mathcal{J}_k\}$ is monotonically nonincreasing. Since it is bounded below, it converges to a finite limit.
\end{IEEEproof}

Next, we consider the optimization of the maximum likelihood problem with the shape parameter \(p\) fixed, which reduces to a factorization problem with an \(\ell_p\)-norm objective.

\section{Optimization on the Quadratic Fitting Problem}
\label{sec:quadratic_fitting_optimization}

We consider the optimization of the quadratic fitting model under a
generalized $p$-power loss. Minimizing $S_p(\Theta)$ is
equivalent to minimizing the aggregate fitting error
\begin{equation}
\label{eq:general_p_fitting}
\min_{f\in\mathcal{F},\,\{\tau_i\}_{i=1}^n}
\quad
F(\Theta):= \frac{1}{p}
\sum_{i=1}^{n}
\|x_i-f(\tau_i)\|_p^p,
\end{equation}
where $x_i\in\mathbb{R}^D$ denotes the observed data, $\tau_i\in
\mathbb{R}^d$ is its latent coordinate, and $\mathcal{F}$ is a
prescribed family of subspace-constrained quadratic functions.

By explicitly substituting the quadratic representation of \(f\) and incorporating the orthogonality constraints, the fitting problem can be reformulated as follows:
\begin{equation}
\label{eq:quadratic_p_optimization}
\begin{aligned}
\min_{c,U,V,\mathcal{A},\,\{\tau_i\}_{i=1}^n}
\quad&\frac{1}{p}
\sum_{i=1}^{n}
\left\|
c+U\tau_i+V\mathcal{A}(\tau_i,\tau_i)-x_i
\right\|_p^p
\\
\text{subject to}\quad&
U^\top U=I_d,\;
V^\top V=I_s,\;
U^\top V=0.
\end{aligned}
\end{equation}

Before presenting the algorithms for solving the nonconvex factorization problem in~\eqref{eq:quadratic_p_optimization}, we first investigate the identifiability of the model, particularly whether distinct parameter configurations can yield the same function. We then derive the corresponding gradients and examine the geometric structure of the orthogonality constraints.

\subsection{Identifiability}
\label{subsec:identifiability}

Identifiability concerns whether the fitted observations uniquely
determine the parameters. In the quadratic manifold model, latent
coordinate changes and equivalent quadratic-map representations
produce non-identifiability.

To make this explicit, let $Q\in\mathbb{R}^{d\times d}$ and
$O\in\mathbb{R}^{s\times s}$ be orthogonal matrices. Consider the
transformation
\begin{equation}
\label{eq:identifiability_transform}
\widetilde U=UQ,\qquad
\widetilde V=VO,\qquad
\widetilde\tau_i=Q^\top\tau_i,
\end{equation}
together with the transformed quadratic map
\begin{equation}
\label{eq:quadratic_map_transform}
\widetilde{\mathcal A}(a,b)
=
O^\top\mathcal A(Qa,Qb).
\end{equation}
Then
\[
\begin{aligned}
\widetilde U\widetilde\tau_i
&=UQQ^\top\tau_i
=U\tau_i,\\
\widetilde V\widetilde{\mathcal A}
(\widetilde\tau_i,\widetilde\tau_i)
&=
VOO^\top\mathcal A(\tau_i,\tau_i)
=V\mathcal A(\tau_i,\tau_i).
\end{aligned}
\]
Hence, the fitted point remains unchanged:
\[
\widetilde f(\widetilde\tau_i)=f(\tau_i).
\]
The orthogonality constraints are also preserved. Thus, the
parameterization is invariant under orthogonal changes of basis in
both the latent tangent space and the quadratic subspace.

This invariance implies that the individual basis matrices and
quadratic coefficients are not uniquely identifiable without
additional conventions. However, the fitted manifold and the
orthogonal projectors
\[
P_U=UU^\top,\qquad P_V=VV^\top,
\]
are invariant under the corresponding transformations and may
therefore be identifiable even when the basis matrices themselves
are not.

\subsection{Closed-Form Expressions for the Gradients}
\label{subsec:sqmf_gradient}

We first derive the Euclidean gradients with respect to the
location parameter, the basis matrices, and the latent coordinates.
For notational convenience, define the residual matrix as
\begin{equation}
\label{eq:sqmf_residual}
R=X-c\mathbf{1}_n^\top-UT-V\Psi
\in\mathbb{R}^{D\times n},
\end{equation}
where $\Psi = [{\cal A}(\tau_1,\tau_1),...,{\cal A}(\tau_n,\tau_n)]\in {\mathbb R}^{s\times n}$.
To simplify the gradient expressions, we introduce the residual gradient matrix for \(p>1\), defined as follows:
\begin{equation}
\label{eq:sqmf_residual_score}
G:=\nabla_R F(\Theta)=|R|^{p-1}\odot {\rm sgn}(R),
\end{equation}
where the absolute value and exponentiation are applied
elementwise, and $\odot$ denotes the Hadamard product.
Equivalently, $G_{ij}=|R_{ij}|^{p-1} {\rm sgn}(R_{ij})$, with the
corresponding score defined to be zero whenever $R_{ij}=0$.
Thus, $G$ is the derivative of $F$ with respect to $R$.
\paragraph{Euclidean Gradient for $c$, $U$ and $V$}
Using the residual definition in \eqref{eq:sqmf_residual},
the gradient with respect to the location parameter is
\begin{equation}
\label{eq:sqmf_grad_c}
\nabla_c F=-G\mathbf{1}_n.
\end{equation}
The Euclidean gradients with respect to the two basis matrices are
\begin{equation}
\label{eq:sqmf_grad_U}
\nabla_U F=-GT^\top,
\end{equation}
and
\begin{equation}
\label{eq:sqmf_grad_V}
\nabla_V F=-G\Psi^\top.
\end{equation}

\paragraph{Gradient for $\mathcal \tau_i$} Here, we derive the gradient with respect to an individual latent
coordinate $\tau_i$. Recall that only the $i$-th column $G_{:,i}$ depends on $\tau_i$. 
The chain rule gives
\begin{equation}
\label{eq:sqmf_grad_tau}
\nabla_{\tau_i}F
=
-U^\top G_{:,i}
-J_{\mathcal{A}}(\tau_i)^\top V^\top G_{:,i},
\end{equation}
where $J_{\mathcal{A}}(\tau_i)\in\mathbb{R}^{s\times d}$
is the Jacobian of the mapping
$\tau\mapsto\mathcal{A}(\tau,\tau)$ evaluated at $\tau_i$.

Because $\mathcal{A}$ is symmetric and bilinear, its directional derivative satisfies
\begin{equation}
\label{eq:sqmf_quadratic_jacobian}
J_{\mathcal{A}}(\tau)[h]
=
\mathcal{A}(h,\tau)+\mathcal{A}(\tau,h)
=
2\mathcal{A}(\tau,h),
\qquad h\in\mathbb{R}^d.
\end{equation}
Consequently, the Jacobian of \(\mathcal A(\tau,\tau)\) with respect to \(\tau\) admits the following closed-form expression:
\[
J_{\mathcal A}(\tau) = 2\mathcal A(\tau) \in {\mathbb R}^{s\times d},
\]
where the $(i,j)$th entry of $\mathcal A(\tau)$ is
$\{\mathcal A(\tau)\}_{i,j} = \sum_{k} {\mathcal A}_{i,j,k}\tau_k.$

\paragraph{Gradient for $\mathcal A$} The gradient with respect to the coefficients of
$\mathcal{A}$ can be derived by differentiating the quadratic
features in $Q$ with respect to those coefficients. In
particular, for a perturbation $\Delta\mathcal{A}$,
the corresponding first-order variation of the residual matrix is
\[
\Delta R
=
-V\Delta \Psi,
\]
where
\[
\Delta \Psi
=
\left[
\Delta\mathcal{A}(\tau_1,\tau_1),\ldots,
\Delta\mathcal{A}(\tau_n,\tau_n)
\right].
\]
Consequently, the directional derivative of the objective is
\begin{equation}
\label{eq:sqmf_directional_A}
D_{\mathcal{A}}F[\Delta\mathcal{A}]
=
-\left\langle G,V\Delta \Psi\right\rangle_{\rm F},
\end{equation}
where $\langle\cdot,\cdot\rangle_{\rm F}$ denotes the Frobenius
inner product. This expression provides the gradient with
respect to any chosen parameterization of the symmetric
bilinear map $\mathcal{A}$.

Let $H=V^\top G\in\mathbb{R}^{s\times n}$. By the chain
rule, the gradient with respect to each tensor entry is
\begin{equation}
\label{eq:sqmf_tensor_gradient_entry}
\begin{aligned}
\frac{\partial F}{\partial\mathcal{A}_{kab}}
&= -\sum_{i=1}^{n} H_{ki}\tau_{ai}\tau_{bi},\\
&\qquad k=1,\ldots,s,\quad a,b=1,\ldots,d.
\end{aligned}
\end{equation}
Equivalently, the tensor gradient admits the compact representation
\begin{equation}
\label{eq:sqmf_tensor_gradient}
\nabla_{\mathcal{A}}F
=
-\sum_{i=1}^{n}
h_i\otimes\tau_i\otimes\tau_i,
\end{equation}
where $h_i=V^\top G_{:,i}\in\mathbb{R}^s$ and $\otimes$
denotes the outer product. The resulting gradient is symmetric
in its last two modes, consistent with the symmetry constraint
on $\mathcal{A}$.

\subsection{Geometry of the Orthogonal Constraints}
\label{subsec:constraint_geometry}

Directly optimizing $U$ and $V$ is challenging because they must jointly
satisfy individual orthonormality and mutual orthogonality constraints.
Notably, these three constraints can be enforced simultaneously by
requiring the concatenated matrix $[U, V]$ to have orthonormal columns.
Accordingly, we introduce the following reparameterization:
$W=[U,V]\in\mathbb{R}^{D\times(d+s)}$. The constraints can therefore be equivalently reformulated as follows:
\[
W^\top W=I_{d+s}.
\]
Consequently, provided that \(d+s\leq D\), the matrix \(W\) belongs to the Stiefel manifold
\[
\operatorname{St}(D,d+s)
=
\left\{
W\in\mathbb{R}^{D\times(d+s)}
:\ W^\top W=I_{d+s}
\right\}.
\]
The tangent space of the Stiefel manifold at \(W\) is given by
\begin{equation}\notag
\begin{aligned}
T_W\operatorname{St}(D,d+s)=
\bigl\{Z\in\mathbb{R}^{D\times(d+s)}:
 W^\top Z+Z^\top W=0\bigr\}.
\end{aligned}
\end{equation}
Given an ambient Euclidean gradient \(G_W\), its orthogonal projection onto the tangent space under the standard Euclidean metric is
\begin{equation}
\label{eq:stiefel_projection}
\operatorname{Proj}_{T_W}(G_W)
=
G_W-W\,\operatorname{sym}(W^\top G_W),
\end{equation}
where
$\operatorname{sym}(M)=\frac{1}{2}(M+M^\top)$

This representation handles all three orthogonality constraints by optimizing the concatenated basis directly on the Stiefel manifold. A Riemannian gradient step can use the projected gradient, followed by a retraction that restores orthonormality.

\subsection{KKT Conditions}
\label{subsec:gradient_kkt}

To derive the constrained first-order conditions, we introduce
symmetric Lagrange multipliers
$\Lambda_U\in\mathbb{R}^{d\times d}$ and
$\Lambda_V\in\mathbb{R}^{s\times s}$, and a multiplier
$M\in\mathbb{R}^{d\times s}$ for the cross-orthogonality constraint.
The Lagrangian is
\begin{align}
\label{eq:lagrangian}
\mathcal L
={}&F
+\frac12\operatorname{tr}
\left[\Lambda_U(U^\top U-I_d)\right]\nonumber\\
&+\frac12\operatorname{tr}
\left[\Lambda_V(V^\top V-I_s)\right]
+\operatorname{tr}(M^\top U^\top V).
\end{align}
The corresponding stationarity conditions for $U$ and $V$ are
\begin{align}
\label{eq:kkt_U}
\nabla_U F+U\Lambda_U+VM^\top&=0,\\
\label{eq:kkt_V}
\nabla_V F+V\Lambda_V+UM&=0.
\end{align}
Stationarity with respect to the unconstrained parameters requires
\begin{equation}
\label{eq:kkt_free}
\nabla_cF=0,\quad
\nabla_{\tau_i}F=0,\quad i=1,\ldots,n,
\quad
\nabla_{\mathcal A}F=0.
\end{equation}
These equations must be accompanied by the primal feasibility
conditions
\begin{equation}
\label{eq:kkt_feasibility}
U^\top U=I_d,\qquad
V^\top V=I_s,\qquad
U^\top V=0.
\end{equation}
Equations \eqref{eq:kkt_U}--\eqref{eq:kkt_feasibility} characterize
the first-order KKT conditions for differentiable objectives,
under the usual constraint qualification. When $p=1$, the gradient
terms are replaced by appropriate subgradients.

Under standard constraint qualifications, the KKT conditions are necessary for local optimality but generally do not guarantee global optimality because the joint fitting problem is nonconvex. 

\section{Algorithms for SQMF}
\label{sec:immediate_residual_gradient}

In this section, we develop a gradient-based optimization algorithm
for the Subspace-Constrained Quadratic Matrix Factorization (SQMF)
model by fixing the shape parameter $p$ of the generalized Gaussian
noise distribution. 

For a fixed $p$, maximizing the likelihood is
equivalent to minimizing the corresponding entrywise $p$-power
residual loss.
\begin{equation}
\label{eq:sqmf_matrix_objective}
\begin{aligned}
\min_{c,U,V,T,\mathcal{A}}
\quad & F(c,U,V,T,\mathcal{A})
=\frac{1}{p} \sum_{ij} |R_{ij}|^p
\\
\text{subject to}\quad &
U^\top U=I_d,\;
V^\top V=I_s,\;
U^\top V=0.
\end{aligned}
\end{equation}

For \(1<p<2\), the resulting optimization problem generally lacks a closed-form solution. We therefore adopt a gradient-based approach to optimize the model parameters while preserving the orthogonality constraints on the basis matrices.

Since the variables \(c\), \(\mathcal A\), and \(\{\tau_i\}\) are unconstrained, they can be updated using gradient descent. The step sizes can be determined by standard line-search strategies, such as backtracking line search, to ensure sufficient decrease in the objective function.

\subsection{Optimization on the Stiefel Manifold}
For the orthonormal basis matrices \(U\) and \(V\), we use Riemannian gradient descent with retraction~\cite{absil2008optimization} to maintain orthogonality. Since \(U\) and \(V\) must be individually orthonormal and mutually orthogonal, we update them jointly.

Let \(H=[\nabla_U F,\nabla_V F]\) denote the Euclidean gradient of \(F\) with respect to \(W\). From~\ref{subsec:constraint_geometry}, we know the Riemannian gradient yields a closed-form representation as
\[
H_{\mathrm{tan}} = \operatorname{Proj}_{T_W}(H)
\]

Given a step size \(\alpha_W>0\), we perform a descent step along the negative Riemannian gradient and then retract the resulting point onto the Stiefel manifold:
\begin{equation}
\label{eq:stiefel_update}
W_{\mathrm{new}}
=
\operatorname{Retr}_{W}
\left(-\alpha_W H_{\mathrm{tan}}\right).
\end{equation}
Here, \(\operatorname{Retr}_{W}\) retracts the updated point $W - \alpha_W H_{\mathrm{tan}}$ onto the Stiefel manifold, preserving orthogonality. A commonly used choice of retraction is based on the QR decomposition. 
The updated basis matrices $U_{\mathrm{new}}$ and
$V_{\mathrm{new}}$ are obtained by partitioning
$W_{\mathrm{new}}$ as follows:
\[
W_{\mathrm{new}}=[U_{\mathrm{new}},V_{\mathrm{new}}].
\]
This joint update automatically preserves the individual orthonormality of \(U\) and \(V\), as well as their mutual orthogonality.
\begin{algorithm}[t]
\caption{Gradient Descent}
\label{alg:sqmf_gd}
\begin{algorithmic}[1]
\REQUIRE Data $X$, dimensions $d,s$, parameter $p>1$, step sizes
$\alpha_c,\alpha_T,\alpha_{\mathcal A},\alpha_W$, tolerance $\varepsilon$, and $K_{\max}$.
\ENSURE $c,U,V,T,\mathcal A$.

\STATE Initialize $c,U,V,T,\mathcal A$ with
$U^\top U=I_d$, $V^\top V=I_s$, and $U^\top V=0$.
\FOR{$k=1,\ldots,K_{\max}$}
    \STATE Compute $Q_{:,i}=\mathcal A(\tau_i,\tau_i)$,
    $R=X-c\mathbf 1_n^\top-UT-V\Psi$,
    $G=|R|^{p-1}\odot {\rm sgn}(R)$,  $H=V^\top G$, and $F^{(k)}=\|R\|_p^p$.
    \STATE Compute gradients:
    \[
    \begin{aligned}
    &g_c=-G\mathbf 1_n, \\
    &g_U=-GT^\top,\\
    &g_V=-G\Psi^\top, \\
    &g_{\tau_i}=-U^\top G_{:,i}-2J_i^\top V^\top G_{:,i},\\
    &(g_{\mathcal A})_{kab}
    =-\sum_i \{V^TG\}_{ki} T_{ai}T_{bi},
    \end{aligned}
    \]
    where $(J_i)_{ka}=\sum_b\mathcal A_{kab}T_{bi}$.
    \STATE Update $c,T,\mathcal A$ by gradient descent using
    $g_c,g_{\tau_i},g_{\mathcal A}$, respectively.
    \STATE Set $W=[U,V]$, $H_W=[g_U,g_V]$, and
    \[
    \begin{aligned}
    W\leftarrow\operatorname{QR}\!\bigl(&W-\alpha_W[H_W\\
    &-W\operatorname{sym}(W^\top H_W)]\bigr),
    \end{aligned}
    \]
    where $\operatorname{sym}(M)=(M+M^\top)/2$.
    \STATE Partition $W=[U,V]$ and recompute $F^{(k+1)}$.
    \IF{$|F^{(k+1)}-F^{(k)}|/\max(1,|F^{(k)}|)<\varepsilon$}
        \STATE \textbf{break}
    \ENDIF
\ENDFOR
\STATE \textbf{return} $c,U,V,T,\mathcal A$.
\end{algorithmic}
\end{algorithm}

\subsection{Gradient Descent and Its Immediate Residual Gradient Update Variant}
\label{subsec:immediate_residual_update}

\begin{algorithm}[!t]
\caption{Immediate Residual Gradient Update (IRGU)}
\label{alg:irgu_sqmf}
\begin{algorithmic}[1]
\REQUIRE $X,p>1$, initial parameters $(c,U,V,T,\mathcal A)$,
step sizes $\alpha_c,\alpha_W,\alpha_T,\alpha_{\mathcal A}$,
tolerance $\varepsilon$, and $K_{\max}$.
\ENSURE Estimated parameters and residual $R$.

\STATE $\Psi\leftarrow[\mathcal A(\tau_i,\tau_i)]_{i=1}^n$,
$R\leftarrow X-c\mathbf 1_n^\top-UT-V\Psi$.

\FOR{$k=1,\ldots,K_{\max}$}
    \STATE $G\leftarrow |R|^{p-1}\odot\operatorname{sgn}(R)$.

    \STATE \textbf{Update $c$:}
    $\Delta c\leftarrow\alpha_cG\mathbf 1_n$;
    $R\leftarrow R-\Delta c\mathbf 1_n^\top$;
    $c\leftarrow c+\Delta c$.

    \STATE \textbf{Update $(U,V)$:}
    $G\leftarrow |R|^{p-1}\odot\operatorname{sgn}(R)$;
    $W\leftarrow[U,V]$, $H\leftarrow[-GT^\top,-G\Psi^\top]$;
    \[
    W_{\rm new}\leftarrow
    \operatorname{Retr}_W\big(
    -\alpha_W[H-W\operatorname{sym}(W^\top H)]
    \big).
    \]
    Partition $W_{\rm new}=[U_{\rm new},V_{\rm new}]$ and update
    \[
    R\leftarrow R-(U_{\rm new}-U)T
                 -(V_{\rm new}-V)\Psi.
    \]
    $(U,V)\leftarrow(U_{\rm new},V_{\rm new})$.

    \STATE \textbf{Update $T$:}
    $G\leftarrow |R|^{p-1}\odot\operatorname{sgn}(R)$;
    \[
    \tau_i^{\rm new}\leftarrow\tau_i-\alpha_T
    \left[-U^\top G_{:,i}
    -J_{\mathcal A}(\tau_i)^\top V^\top G_{:,i}\right].
    \]
    Compute $\Psi_{\rm new}=[\mathcal A(\tau_i^{\rm new},
    \tau_i^{\rm new})]_{i=1}^n$ and update
    \[
    R\leftarrow R-U(T_{\rm new}-T)-V(\Psi_{\rm new}-\Psi).
    \]
    $(T,\Psi)\leftarrow(T_{\rm new},\Psi_{\rm new})$.

    \STATE \textbf{Update $\mathcal A$:}
    $G\leftarrow |R|^{p-1}\odot\operatorname{sgn}(R)$;
    $\mathcal A_{\rm new}\leftarrow
    \mathcal A-\alpha_{\mathcal A}\nabla_{\mathcal A}F$;
    compute $\Psi_{\rm new}$ and update
    \[
    \begin{aligned}
    R&\leftarrow R-V(\Psi_{\rm new}-\Psi),\\
    (\mathcal A,\Psi)&\leftarrow
    (\mathcal A_{\rm new},\Psi_{\rm new}).
    \end{aligned}
    \]

    \STATE $F_k\leftarrow\frac{1}{p}\|R\|_p^p$.
    \IF{the stopping criterion is satisfied}
        \STATE \textbf{break}
    \ENDIF
\ENDFOR
\RETURN $(c,U,V,T,\mathcal A,R)$.
\end{algorithmic}
\end{algorithm}

We first present the standard gradient descent (GD) algorithm in Algorithm~\ref{alg:sqmf_gd}, which sequentially updates \(c\), \(\{\tau_i\}\), and \(\mathcal{A}\). Note that each parameter update modifies the residual matrix \(R\), thereby changing the gradients associated with the subsequent updates. To account for these changes, we further introduce an {\it immediate residual update} strategy, which updates \(R\) and recomputes the gradient matrix \(G\) immediately after each parameter update. Consequently, each subsequent gradient is evaluated using the most up-to-date residual information. This strategy is conceptually related to sequential residual methods for solving nonlinear systems, where residuals are updated immediately as the solution is progressively refined; see, e.g., \cite{birgin2022secant}.


The key idea of the immediate residual update strategy is to refresh the residual matrix immediately after each variable update and use the updated residuals to compute the gradients for subsequent updates. This incremental gradient updating mechanism is closely related to the approaches studied in~\cite{tseng2014incrementally}. The details of the proposed strategy are presented below.

\paragraph{Update of the center $c$}
Let $\Delta c=c_{\mathrm{new}}-c$. The residual matrix is updated as
\begin{equation}
\label{eq:residual_update_c}
R\leftarrow R-\Delta c\,\mathbf{1}_n^\top.
\end{equation}

\paragraph{Update $U$ and $V$}
When $U$ and $V$ are updated simultaneously while $T$ and $\Psi$
remain fixed, the residual update becomes
\begin{equation}
\label{eq:residual_joint_update}
R\leftarrow
R-(U_{\mathrm{new}}-U)T
 -(V_{\mathrm{new}}-V)\Psi.
\end{equation}

\paragraph{Update of the latent coordinates $T$}
Updating $T$ also changes the quadratic feature matrix $\Psi$. Specifically, define
\[
\Psi_{\mathrm{new}}
=
\left[
\mathcal{A}(\tau_{1,\mathrm{new}},\tau_{1,\mathrm{new}}),
\ldots,
\mathcal{A}(\tau_{n,\mathrm{new}},\tau_{n,\mathrm{new}})
\right].
\]
The corresponding residual update is
\begin{equation}
\label{eq:residual_update_T}
R\leftarrow
R-U(T_{\mathrm{new}}-T)
-V(\Psi_{\mathrm{new}}-\Psi).
\end{equation}

\paragraph{Update of the quadratic map $\mathcal{A}$}
When $\mathcal{A}$ changes while $T$ remains fixed, we have the residual matrix is updated as
\begin{equation}
\label{eq:residual_update_A}
R\leftarrow
R-V\left[
\Delta\mathcal{A}(\tau_1,\tau_1),\ldots,
\Delta\mathcal{A}(\tau_n,\tau_n)
\right],
\end{equation}
where
$\Delta\mathcal{A}=\mathcal{A}_{\mathrm{new}}-\mathcal{A}$.

After each accepted parameter update, we recompute the score matrix $G=|R|^{p-1}\odot {\rm sgn}(R)$,
where the power and product are taken elementwise. The updated $G$
is then used to evaluate the gradient of the next parameter block.
This strategy ensures that each gradient evaluation is consistent
with the latest residual matrix while avoiding unnecessary
recomputation of $R$ from the original data. Algorithm~\ref{alg:irgu_sqmf} summarizes the proposed immediate residual gradient update procedure.

For both standard gradient descent and its immediate residual gradient update variant, we employ a backtracking line-search strategy~\cite{nocedal2006numerical} to determine an appropriate step size. This strategy ensures a sufficient decrease in the objective function at each iteration while avoiding excessively small updates, thereby promoting stable and efficient convergence.


\section{Numerical Experiments}
\label{sec:numerical_experiments}

This section evaluates the proposed alternating maximum likelihood
estimation framework for subspace-constrained quadratic matrix
factorization (AMLE-SQMF) from four perspectives: optimization
efficiency, parameter estimation accuracy, the effects of ambient
dimension and sample size, and manifold reconstruction performance.
The implementation is publicly available on GitHub.\footnote{\url{https://github.com/liangzhuan/AMLE-SQMF}}
The experiments are designed to address
the following questions:
\begin{itemize}
    \item Does the immediate residual gradient update (IRGU) improve the computational efficiency of the geometric fitting subproblem?
     \item How do the ambient dimension and the number of observations
    affect the estimation accuracy of the shape parameter?
    \item How accurately can AMLE-SQMF estimate the Generalized Gaussian shape parameter $p$ and scale
    parameter $\alpha$ under different noise levels and neighborhood sizes?
    \item Does adapting the residual model improve manifold
    reconstruction?
\end{itemize}

\subsection{Speed Comparison: IRGU vs. GD}
\label{sec:expA_sphere}

We compare the immediate residual gradient update (IRGU) in
Algorithm~\ref{alg:irgu_sqmf} with gradient descent (GD) in
Algorithm~\ref{alg:sqmf_gd}. Both methods minimize the objective $F$
in~\eqref{eq:sqmf_matrix_objective}. Prescribing the fitting exponent
isolates the geometric optimization from noise-parameter estimation.
The known sphere geometry allows the local fit to be inspected,
while objective trajectories and optimization times compare the
efficiency of the two solvers.

We uniformly sample 1,000 points from the unit sphere in
$\mathbb{R}^3$ and add independent generalized Gaussian noise to each coordinate. The noise shape and fitting exponent are matched, with $p\in\{1.25,1.50,1.75,2.00\}$, and the noise standard deviation is fixed at $\sigma=0.05$ to control noise intensity. For a prespecified noisy point, its 200 nearest noisy neighbors form the fitting neighborhood.

\begin{figure}[!t]
\centering
\includegraphics[width=0.6\columnwidth]{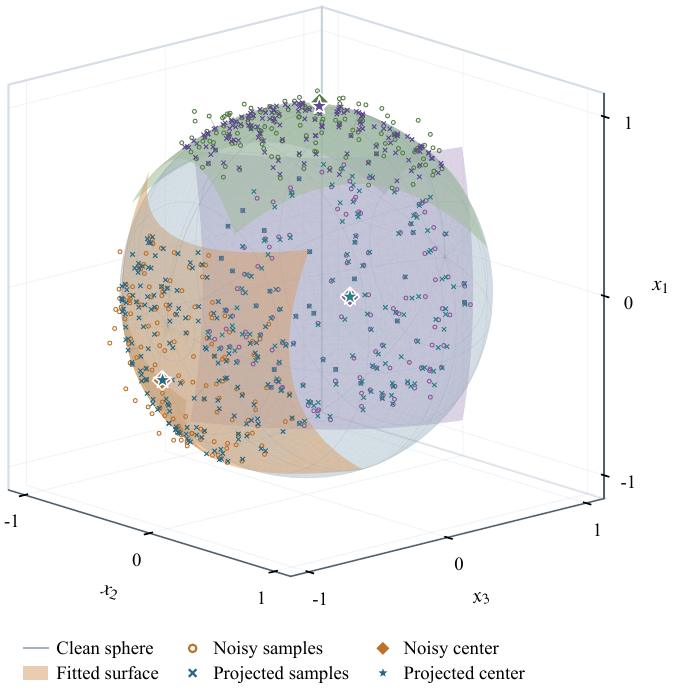}
\caption{Illustration of local quadratic surface fitting to noisy sphere samples using IRGU. The blue sphere shows the original unit sphere, and the colored patches show the fitted quadratic surfaces. Circles and crosses represent noisy samples and their Euclidean projections onto the fitted surfaces, respectively. Diamonds and
stars represent the noisy neighborhood centers and their
projections, respectively.}
\label{fig:expA_sphere}
\end{figure}

\begin{figure}[!t]
\centering
\includegraphics[width=\columnwidth]{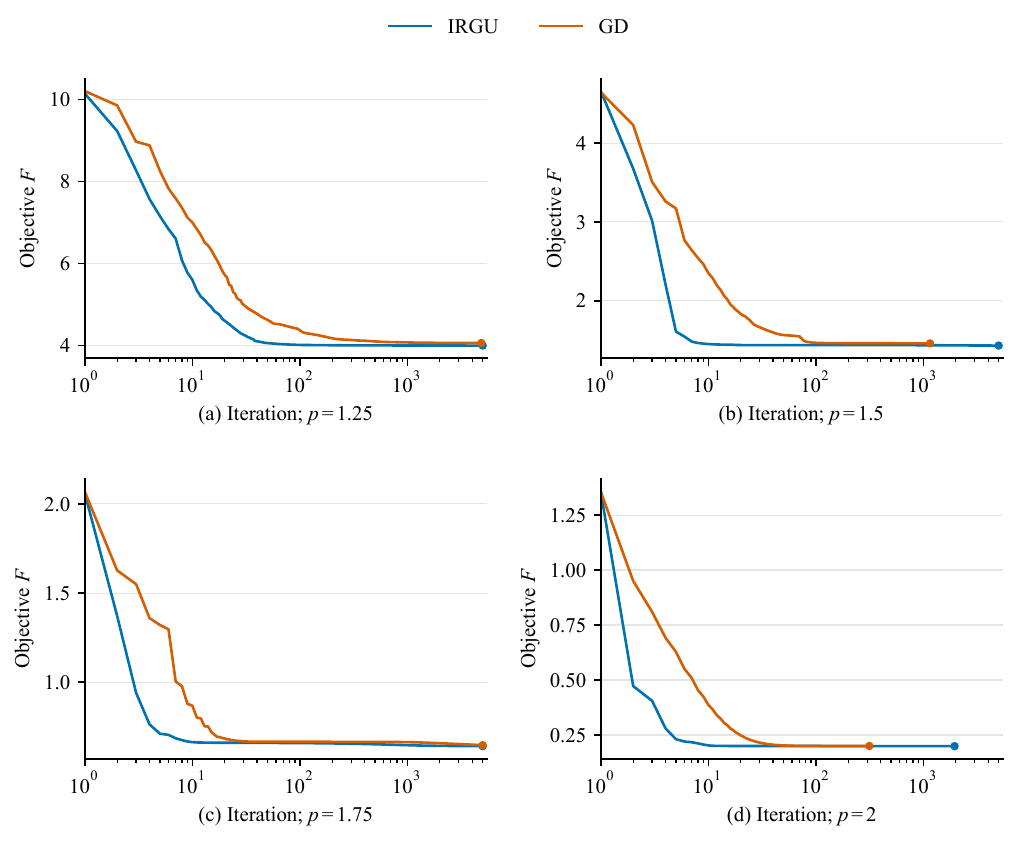}
\caption{Objective function value $F$ versus iteration number for
GD and IRGU when fitting a local quadratic surface to noisy sphere
samples. Panels (a)--(d) use $p=1.25,1.5,1.75,2$, respectively,
for both the noise shape and the fitting loss exponent.}
\label{fig:expA_iterations}
\end{figure}

\begin{table}[!htbp]
\centering
\caption{Optimization time (seconds) on the sphere example.}
\label{tab:expA_time}
\footnotesize
\renewcommand{\arraystretch}{0.8}
\begin{tabular*}{\columnwidth}{@{\extracolsep{\fill}}lcccc@{}}
\toprule
Methods / $p$ & 1.25 & 1.50 & 1.75 & 2.00 \\
\midrule
GD   & 13.91 & 3.01 & 12.41 & 0.81 \\
IRGU &  2.32 & 2.24 &  2.42 & 0.75 \\
\bottomrule
\end{tabular*}
\end{table}

Fig.~\ref{fig:expA_iterations} shows that both methods converge to
comparable objective values, while IRGU reduces the objective more
rapidly during the early iterations.
Table~\ref{tab:expA_time} reports the optimization time of each run.
IRGU requires less time in all four conditions, with the largest
reductions at $p=1.25$ and $p=1.75$.
Fig.~\ref{fig:expA_sphere} further illustrates how the fitted quadratic
surface captures the local sphere curvature, together with the noisy
samples and their projections.

To assess the local fitting performance across different neighborhood centers, we repeat the local
fitting experiment at 30 randomly selected points for each
exponent. Let $x_q\in\mathbb{R}^3$ denote the noisy observation used
as the neighborhood center, and let $\widehat f_q$ denote the fitted
quadratic map. We compute the minimum Euclidean distance from each
neighborhood center to its fitted quadratic surface as
$
\min_{\tau}\|x_q-\widehat{f}_q(\tau)\|_2$.

Table~\ref{tab:expA_delta} reports the mean and sample standard deviation
of these distances over the 30 neighborhood centers
for each exponent.

\begin{table}[!htbp]
\centering
\caption{Mean and sample standard deviation of the
minimum Euclidean distances from 30 noisy neighborhood
centers to their fitted quadratic surfaces for each $p$.}
\label{tab:expA_delta}
\begingroup
\footnotesize
\renewcommand{\arraystretch}{0.8}
\begin{tabular*}{\columnwidth}{@{\extracolsep{\fill}}lcccc@{}}
\toprule
$p$ & 1.25 & 1.50 & 1.75 & 2.00 \\
\midrule
Mean & 0.046 & 0.032 & 0.033 & 0.036 \\
SD   & 0.038 & 0.027 & 0.029 & 0.038 \\
\bottomrule
\end{tabular*}
\endgroup
\end{table}
{ We observe that both algorithms eventually achieve the same optimal objective value. However, by updating the gradient more frequently through immediate residual updates, IRGU converges faster than standard GD, requiring fewer outer iterations to reach the optimum.}

\subsection{Effect of Ambient Dimension and Sample Size}

Here, we examine how ambient dimension and sample size affect
shape-parameter estimation through an experiment on synthetic data. We generate synthetic observations
from a two-dimensional quadratic manifold embedded in
$\mathbb{R}^D$:
\[
x_i = c + U\tau_i + V{\cal A}(\tau_i,\tau_i) + \epsilon_i,
\]
where $\tau_i\in\mathbb{R}^2$ is the latent coordinate, and the
noise entries independently follow a generalized Gaussian
distribution with shape $p^*=1.4$ and scale $\alpha^*=0.02$.
We consider $D\in\{30,60,90,120,150\}$ and
$n\in\{1000,1500,2000,2500,3000\}$, yielding 25 configurations,
each evaluated over 50 repetitions. 
By varying one factor while
holding the other fixed, we assess their individual effects using
absolute empirical bias, standard deviation (SD), and
mean squared error (MSE).

\begin{figure}[!t]
\centering
\includegraphics[
    width=\columnwidth,
    trim=0 8.5bp 0 0,
    clip
]{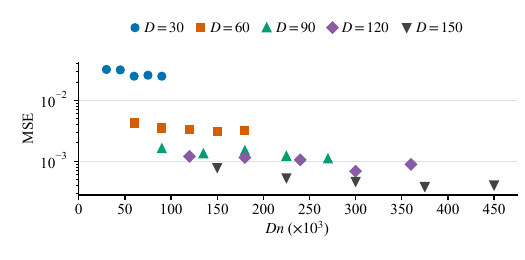}
\caption{Shape-estimation MSE across the $(D,n)$ configurations.
 }
\label{fig:expB-Dn-mse}
\end{figure}

Fig.~\ref{fig:expB-Dn-mse} shows shape-estimation MSE for \(\widehat{p}\) versus sample size \(n\) and ambient dimension \(D\). MSE generally decreases as \(Dn\) increases, indicating more accurate \(\widehat{p}\) estimates with larger sample sizes and ambient dimensions.

To isolate the effects of \(n\) and \(D\), we examine absolute bias \(|\widehat{p}-p^*|\) by (i) fixing \(D\) and varying \(n\), and (ii) fixing \(n\) and varying \(D\). Increasing \(D\) reduces bias more than increasing \(n\), suggesting a stronger effect of ambient dimension on \(p\) estimation under these settings.


\begin{figure}[!t]
\centering
\includegraphics[width=\columnwidth]
{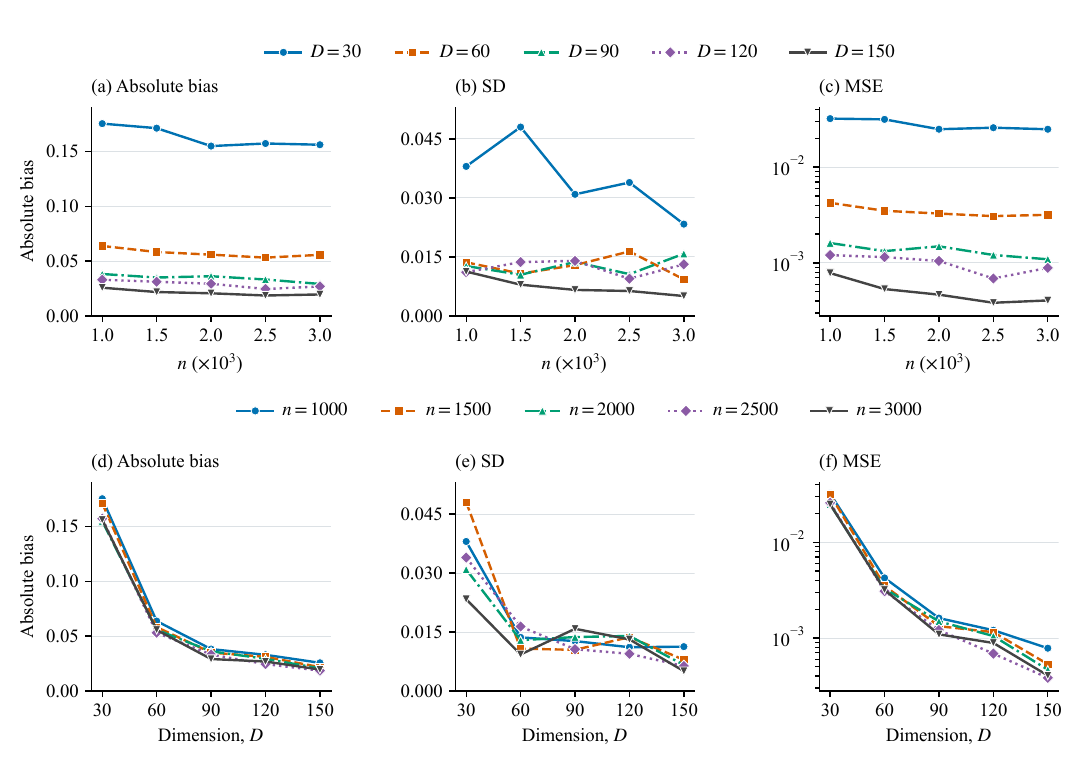}
\caption{Effects of sample size and ambient dimension on
shape-parameter estimation. The top row shows the effects
of varying sample size $n$ at fixed ambient dimension $D$,
whereas the bottom row shows the effects of varying ambient
dimension $D$ at fixed sample size $n$. }
\label{fig:expB-Dn-effects}
\end{figure}

\subsection{Factors Affecting Parameter Estimation}
\label{subsec:expB-parameter-estimation}

Here, we examine how noise scale and neighborhood sample size
affect the joint estimation of $p$ and $\alpha$ on Frey Face~\cite{freyFaceDataset}.
We use generating shapes $p_\star\in\{1.00,1.25,1.50,1.75,2.00\}$
and vary one factor at a time.

\subsubsection{Effect of Noise Scale}
To investigate the effect of the noise scale on the estimation accuracy of \(p\), we conduct experiments using 30 randomly selected images. Specifically, we consider four noise scale settings, $\alpha_\star\in\{0.05,0.10,0.15,0.20\}$, where $\alpha_\star$ controls the magnitude of the additive noise. For each setting, we perform 100 independent trials, with the noise independently resampled in each trial. The noisy observations are generated according to $x_i=x_i^\star+\epsilon_i$,
where $\epsilon_i\sim \operatorname{GGD}(\alpha_\star,p_\star)$.

\begin{figure}[!t]
\centering
\includegraphics[width=\columnwidth]{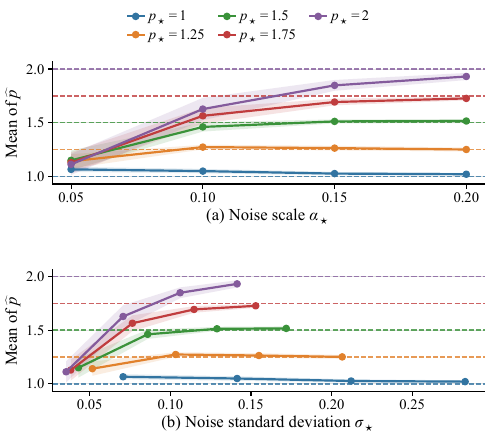}
\caption{Mean estimates of $p$ for 30 Frey Face images versus
(a) the noise scale $\alpha_\star$ and (b) the corresponding
noise standard deviation. Shaded regions indicate one sample
standard deviation; dashed lines indicate the true values of $p$.}
\label{fig:expB-scale-sensitivity}
\end{figure}

\begin{figure}[!t]
\centering
\includegraphics[width=0.98\columnwidth]{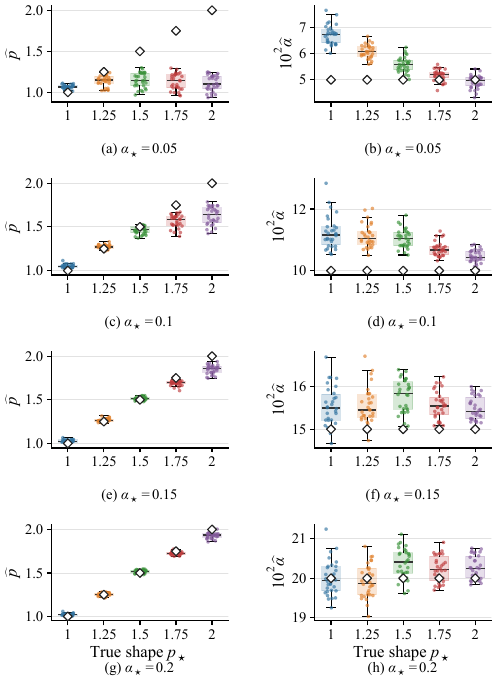}
\caption{Shape (left) and scale (right) estimates over 30 repeats
at different generating scales. Dots show individual estimates;
boxes show the interquartile ranges with medians, and diamonds
indicate the generating parameters. Scale estimates are multiplied
by $10^2$.}
\label{fig:expB-scale-distribution}
\end{figure}

We investigate the effect of the noise scale on the estimation of \(p\) by examining the empirical mean, standard deviation, and distribution of \(\widehat{p}\). We consider both the noise scale \(\alpha_\star\) and its corresponding standard deviation \(\sigma_\star\), which are related by $\sigma_\star=\alpha_\star\sqrt{\frac{\Gamma(3/p_\star)}{\Gamma(1/p_\star)}}$~\cite{nadarajah2005generalized}.
The results are reported in Table~\ref{tab:expB-shape-scale}, Fig.~\ref{fig:expB-scale-sensitivity} and Fig.~\ref{fig:expB-scale-distribution}.

We observe that the estimation accuracy and stability generally improve with increasing noise levels. This is because larger noise scales make the generalized Gaussian noise more dominant relative to the approximation error between \(\widehat{f}\) and the underlying fitting curve, making the residual distribution better match the assumed generalized Gaussian model.

\begin{table}[t]
\centering
\caption{Estimates of \(p\) and \(\alpha\) across 30 Frey Face images under different noise scales. Entries are reported as means \(\pm\) sample standard deviations.}
\label{tab:expB-shape-scale}
\begingroup
\footnotesize
\setlength{\tabcolsep}{0.8pt}
\renewcommand{\arraystretch}{0.8}

\begin{tabular*}{\columnwidth}{@{\extracolsep{\fill}}ccccc@{}}
\toprule
$p_\star$
& $\alpha_\star$ = 0.05
& $\alpha_\star$ = 0.10
& $\alpha_\star$ = 0.15
& $\alpha_\star$ = 0.20 \\
\midrule
\multicolumn{5}{c}{Shape $\widehat p$} \\
1.00
& 1.066 $\pm$ 0.024
& 1.050 $\pm$ 0.020
& 1.027 $\pm$ 0.015
& 1.021 $\pm$ 0.012 \\
1.25
& 1.142 $\pm$ 0.066
& 1.273 $\pm$ 0.021
& 1.264 $\pm$ 0.021
& 1.251 $\pm$ 0.018 \\
1.50
& 1.150 $\pm$ 0.092
& 1.461 $\pm$ 0.043
& 1.513 $\pm$ 0.019
& 1.516 $\pm$ 0.018 \\
1.75
& 1.129 $\pm$ 0.098
& 1.564 $\pm$ 0.073
& 1.693 $\pm$ 0.031
& 1.726 $\pm$ 0.019 \\
2.00
& 1.112 $\pm$ 0.098
& 1.627 $\pm$ 0.104
& 1.848 $\pm$ 0.050
& 1.930 $\pm$ 0.033 \\
\midrule
\multicolumn{5}{c}{Scale $\widehat\alpha$} \\
1.00
& 0.067 $\pm$ 0.004
& 0.112 $\pm$ 0.005
& 0.155 $\pm$ 0.005
& 0.200 $\pm$ 0.004 \\
1.25
& 0.061 $\pm$ 0.003
& 0.111 $\pm$ 0.004
& 0.156 $\pm$ 0.004
& 0.199 $\pm$ 0.004 \\
1.50
& 0.056 $\pm$ 0.003
& 0.110 $\pm$ 0.003
& 0.158 $\pm$ 0.004
& 0.204 $\pm$ 0.004 \\
1.75
& 0.052 $\pm$ 0.002
& 0.107 $\pm$ 0.002
& 0.156 $\pm$ 0.003
& 0.202 $\pm$ 0.003 \\
2.00
& 0.049 $\pm$ 0.003
& 0.105 $\pm$ 0.002
& 0.155 $\pm$ 0.003
& 0.203 $\pm$ 0.003 \\
\bottomrule
\end{tabular*}
\endgroup
\end{table}

\begin{figure}[!t]
\centering
\includegraphics[width=\columnwidth]{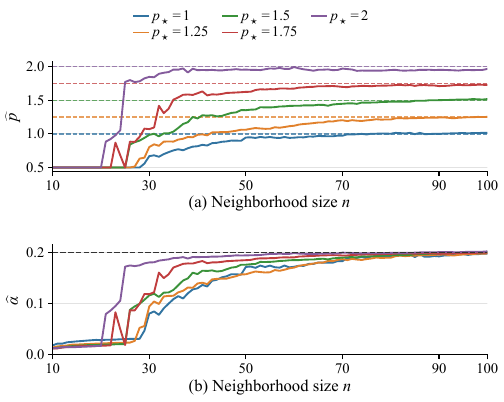}
\caption{Parameter estimates for one image as the nested
neighborhood grows: (a) shape and (b) scale.
}
\label{fig:expB-sample-trajectory}
\end{figure}

\begin{table}[!t]
\centering
\caption{Estimates of \(p\) and \(\alpha\) across 20 Frey Face neighborhood centers with varying sample sizes, where the true noise scale is fixed at \(\alpha_\star=0.2\). Entries are reported as the mean \(\pm\) sample standard deviation.}
\label{tab:expB-sample-size}
\begingroup
\footnotesize
\setlength{\tabcolsep}{0.8pt}
\renewcommand{\arraystretch}{0.8}

\begin{tabular*}{\columnwidth}{@{\extracolsep{\fill}}cccc@{}}
\toprule
$p_\star$ & $n$ = 30 & $n$ = 60 & $n$ = 90 \\
\midrule
\multicolumn{4}{c}{Shape $\widehat p$} \\
1.00
& 0.583 $\pm$ 0.066
& 0.940 $\pm$ 0.016
& 1.007 $\pm$ 0.012 \\
1.25
& 0.817 $\pm$ 0.058
& 1.135 $\pm$ 0.019
& 1.235 $\pm$ 0.019 \\
1.50
& 0.981 $\pm$ 0.029
& 1.450 $\pm$ 0.041
& 1.506 $\pm$ 0.022 \\
1.75
& 1.130 $\pm$ 0.188
& 1.695 $\pm$ 0.030
& 1.723 $\pm$ 0.021 \\
2.00
& 1.748 $\pm$ 0.254
& 1.911 $\pm$ 0.029
& 1.930 $\pm$ 0.027 \\
\midrule
\multicolumn{4}{c}{Scale $\widehat\alpha$} \\
1.00
& 0.055 $\pm$ 0.020
& 0.172 $\pm$ 0.006
& 0.195 $\pm$ 0.004 \\
1.25
& 0.099 $\pm$ 0.014
& 0.174 $\pm$ 0.004
& 0.195 $\pm$ 0.004 \\
1.50
& 0.117 $\pm$ 0.005
& 0.191 $\pm$ 0.006
& 0.201 $\pm$ 0.004 \\
1.75
& 0.126 $\pm$ 0.022
& 0.195 $\pm$ 0.004
& 0.201 $\pm$ 0.004 \\
2.00
& 0.174 $\pm$ 0.021
& 0.196 $\pm$ 0.003
& 0.201 $\pm$ 0.003 \\
\bottomrule
\end{tabular*}
\endgroup
\end{table}

\begin{table*}[!t]
\centering
\caption{Reconstruction MSE for 30 images from the Isomap Face dataset.
Entries are reported as the means $\pm$ sample standard deviations.}
\label{tab:isomap_face_mse}
\begingroup
\renewcommand{\arraystretch}{0.8}
\setlength{\tabcolsep}{7pt}
\footnotesize

\begin{tabular*}{\textwidth}{@{\extracolsep{\fill}}lccccc@{}}
\toprule
Method
& GGD $p_\star$ = 1.25
& GGD $p_\star$ = 1.50
& GGD $p_\star$ = 1.75
& Logistic
& Student-$t$\textsubscript{10} \\
\midrule
AMLE-SQMF (Ours)
& 8.902 $\pm$ 2.226
& \textbf{8.595} $\pm$ 1.949
& \textbf{9.038} $\pm$ 2.042
& \textbf{8.828} $\pm$ 1.901
& \textbf{8.967} $\pm$ 1.989 \\
SQMF ($p$ = 1)
& 9.857 $\pm$ 2.465
& 10.443 $\pm$ 4.573
& 10.247 $\pm$ 1.971
& 9.642 $\pm$ 1.676
& 9.756 $\pm$ 1.551 \\
SQMF ($p$ = 2)
& \textbf{8.895} $\pm$ 2.007
& 9.033 $\pm$ 2.756
& 9.066 $\pm$ 1.923
& 8.884 $\pm$ 1.884
& 8.998 $\pm$ 1.928 \\
LPCA
& 12.121 $\pm$ 2.637
& 12.049 $\pm$ 2.333
& 12.163 $\pm$ 2.532
& 11.815 $\pm$ 2.394
& 11.914 $\pm$ 2.299 \\
SPH
& 14.889 $\pm$ 5.137
& 14.457 $\pm$ 4.230
& 14.565 $\pm$ 4.823
& 14.583 $\pm$ 4.272
& 14.485 $\pm$ 3.832 \\
Q-LS
& 11.923 $\pm$ 2.399
& 11.825 $\pm$ 2.357
& 11.816 $\pm$ 2.251
& 11.614 $\pm$ 2.162
& 11.660 $\pm$ 2.137 \\
MFIT
& 12.382 $\pm$ 2.675
& 11.484 $\pm$ 2.588
& 11.940 $\pm$ 2.043
& 11.755 $\pm$ 2.153
& 11.866 $\pm$ 2.424 \\
LOG-KDE
& 11.670 $\pm$ 2.385
& 11.645 $\pm$ 2.215
& 11.571 $\pm$ 2.179
& 11.501 $\pm$ 2.277
& 11.560 $\pm$ 2.197 \\
\bottomrule
\end{tabular*}
\endgroup
\end{table*}

\begin{figure*}[!t]
\centering
\includegraphics[width=\textwidth]{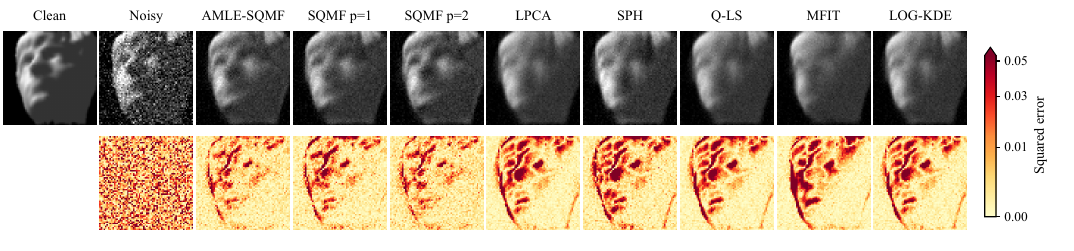}
\caption{Isomap Face reconstructions (top) and pixelwise squared
errors (bottom) under generalized Gaussian noise with $p_\star=1.25$.}
\label{fig:isomap_face_example}
\end{figure*}

\begin{figure}[!t]
\centering
\includegraphics[width=\columnwidth]{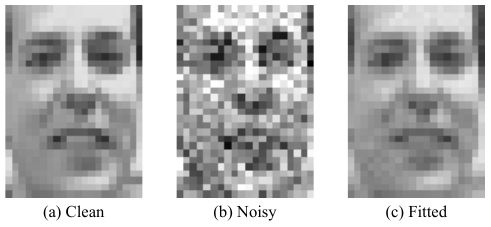}
\caption{Frey Face reconstruction with
$\alpha_\star=0.20$ and $p_\star=2$:
(a) clean image, (b) noisy image, and (c) fitted image
using the estimated shape parameter.}
\label{fig:expB-face-example}
\end{figure}

\subsubsection{Effect of Neighborhood Sample Size}

We investigate the effect of sample size on the estimation of \(p\), with the noise scale fixed at \(\alpha_\star=0.2\). Specifically, we randomly select 20 Frey Face images as neighborhood centers and construct neighborhoods of varying sizes. All images within each neighborhood are then used as input to the maximum likelihood estimation algorithm.

We first consider three sample sizes, \(n\in\{30,60,90\}\), using the same 20 neighborhood centers. We then continuously increase \(n\) from 10 to 100 to examine how the estimated parameter \(\widehat{p}\) approaches its ground-truth value \(p_\star\) as the sample size grows. The results are presented in Table~\ref{tab:expB-sample-size} and Fig.~\ref{fig:expB-sample-trajectory}. Representative image reconstruction results are also provided in Fig.~\ref{fig:expB-face-example}. 

We observe that the estimation accuracy generally improves with increasing sample size, which is consistent with the asymptotic properties of maximum likelihood estimation.

\subsection{Adaptive Manifold Fitting on Face Images}
\label{subsec:isomap_face_fitting}
Here, we examine whether adapting the residual distribution improves manifold reconstruction under different noise conditions. Beyond recovering the noise parameters, this experiment validates whether their estimation benefits the reconstruction.

We use noisy images as inputs to the different algorithms. The clean images consist of 30 randomly selected samples from the Isomap Face dataset~\cite{tenenbaum2000isomap}. We corrupt these images with noise drawn from three different distributions: Gaussian, logistic, and Student's \(t\)-distributions. Reconstruction accuracy is evaluated using the pixelwise mean squared error (MSE). For AMLE-SQMF, we use the estimated shape parameter of the residual distribution to determine the loss function for manifold fitting.

We compare our method with five representative manifold denoising methods under identical noise settings: LPCA~\cite{lpca}, SPH~\cite{li2022efficient}, Q-LS~\cite{sober2020manifold}, MFIT~\cite{fefferman2018fitting}, and LOG-KDE~\cite{ozertem2011locally}. We repeat each experiment multiple times and report the mean squared error (MSE) and standard deviation (SD) in Table~\ref{tab:isomap_face_mse}.

We observe that AMLE-SQMF achieves the lowest reconstruction error in most tested conditions and
outperforms the geometric baselines across all noise distributions. These results support the effectiveness of adaptive quadratic fitting for high-dimensional face reconstruction, especially for logistic and Student-$t$ noise distributions.

Compared with fixed-exponent SQMF, AMLE-SQMF maintains competitive reconstruction accuracy without prescribing the residual exponent. The favorable results under logistic and Student-$t$ noise further show that the usefulness of adaptive residual modeling extends beyond generalized Gaussian perturbations. Thus, the estimated residual model provides an effective basis for manifold reconstruction across the tested noise conditions.

To further illustrate the superior reconstruction performance of AMLE-SQMF, we visualize the reconstructed images and their corresponding error maps in Fig.~\ref{fig:isomap_face_example}. We observe that AMLE-SQMF produces noticeably smaller reconstruction errors than the competing methods.

\subsection{Latent-Space Interpolation of Face Images}
\label{subsec:latent_face_interpolation}

Here, we examine whether the quadratic mapping used by AMLE-SQMF
provides an interpretable representation for image interpolation.
We select 40 images of one subject from Yale Face Database
B~\cite{georghiades2001yale}, with a fixed frontal pose and varying
illumination, and learn the mapping with latent dimension $d=2$.

\begin{figure}[!t]
\centering
\includegraphics[width=\columnwidth]
{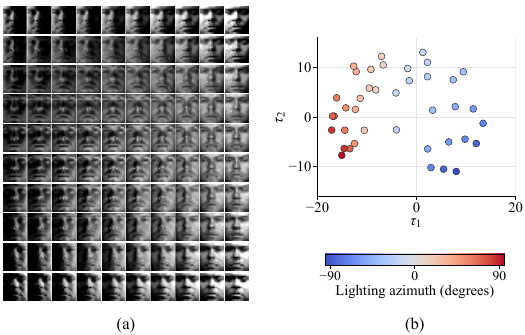}
\caption{Latent-space visualization for 40 images of one subject.
(a) Images generated on a $10\times10$ latent grid.
(b) Fitted coordinates; blue and red indicate negative and positive
lighting azimuths, respectively.}
\label{fig:yale_latent_interpolation}
\end{figure}

The learned map $\widehat f:\mathbb{R}^2\rightarrow\mathbb{R}^D$
allows new images to be generated by sampling latent coordinates.
We uniformly sample a $10\times10$ grid over the fitted coordinate
ranges and map these points back to image space, as shown in
Fig.~\ref{fig:yale_latent_interpolation}(a).
Fig.~\ref{fig:yale_latent_interpolation}(b) shows the fitted latent coordinates of the 40 images, colored by lighting azimuth. The generated images
exhibit gradual illumination changes, while the observed samples
occupy different latent regions according to lighting direction.
These results show that the learned representation supports
interpolation of facial appearance through a compact,
two-dimensional coordinate system.

\section{Conclusion}

In this paper, we proposed an alternating maximum likelihood estimation framework for shape-adaptive subspace-constrained quadratic matrix factorization. Modeling residuals with a generalized Gaussian distribution, the method jointly estimates the quadratic manifold parameters and shape parameter, allowing the fitting objective to adapt to different residual distributions. We derived a profile negative log-likelihood by eliminating the scale parameter through its closed-form maximum likelihood estimator and developed an alternating optimization scheme with convergence analysis. To solve the geometric fitting problem efficiently, we formulated the basis orthogonality constraints on the Stiefel manifold and developed a Riemannian gradient descent algorithm with an immediate residual gradient update (IRGU) strategy. Numerical experiments demonstrated that IRGU improves computational efficiency over conventional gradient descent. They also showed that shape parameter estimation benefits from increased ambient dimension and sample size, while AMLE-SQMF achieves competitive reconstruction performance under generalized Gaussian, logistic, and Student-$t$ noise.

Future work will investigate extensions to more general noise models, including structured and dependent residuals, and higher-order local manifold representations. Another direction is to establish stronger theoretical guarantees for the statistical consistency of the shape parameter estimator and convergence of the alternating optimization algorithm. These extensions may broaden the applicability of adaptive quadratic manifold fitting to complex high-dimensional data.
\par\label{sec:numerical_experiments_end}
\bibliographystyle{IEEEtran}
\bibliography{sample}
\vspace{-4mm}
\begin{IEEEbiographynophoto}{Zheng Zhai}
is currently a faculty member in the Department of Statistics, Faculty of Arts and Sciences at Beijing Normal University, Zhuhai.
He received his Ph.D. degree from the School of Mathematical Sciences at Zhejiang University and worked as a postdoctoral researcher at the University of Toronto and the National University of Singapore. His current research interests include diffusion model, matrix computation, and manifold learning.
\end{IEEEbiographynophoto}
\vspace{-4mm}
\begin{IEEEbiographynophoto}{Zhuan Liang}
 is currently a graduate student in the Department of Statistics, Faculty of Arts and Sciences at Beijing Normal University, Zhuhai.
He received his B.S. degree from the School of Mathematics and Statistics at Shandong University. His current research interests include diffusion models and matrix computation.
\end{IEEEbiographynophoto}
\vspace{-4mm}
\begin{IEEEbiographynophoto}{Sujun Guo}
is currently a graduate student in the Department of Statistics, Faculty of Arts and Sciences, Beijing Normal University, Zhuhai, China. She received the B.S. degree from the School of Mathematics and Statistics, Northeast Normal University. Her research interests include statistical methods and their applications in machine learning.
\end{IEEEbiographynophoto}

\end{document}